# THE FUZZY GENE FILTER: A CLASSIFIER PERFORMANCE ASSESMENT

Meir Perez, Tshilidzi Marwala
University of Johannesburg, School of Electrical and Electronics Engineering
South Africa
mperez@uj.ac.za

**ABSTRACT**

The Fuzzy Gene Filter (FGF) is an optimised Fuzzy Inference System designed to rank genes in order of differential expression, based on expression data generated in a microarray experiment. This paper examines the effectiveness of the FGF for feature selection using various classification architectures. The FGF is compared to three of the most common gene ranking algorithms: t-test, Wilcoxon test and ROC curve analysis. Four classification schemes are used to compare the performance of the FGF vis-à-vis the standard approaches: K Nearest Neighbour (KNN), Support Vector Machine (SVM), Naïve Bayesian Classifier (NBC) and Artificial Neural Network (ANN). A nested stratified Leave-One-Out Cross Validation scheme is used to identify the optimal number top ranking genes, as well as the optimal classifier parameters. Two microarray data sets are used for the comparison: a prostate cancer data set and a lymphoma data set.

Genes ranked by the FGF attained significantly higher accuracies for all of the classifiers tested, on both data sets ($p < 0.0231$ for the prostate data set and $p < 0.1888$ for the lymphoma data set). When using the prostate data set, the FGF performed best on the KNN classifier, achieving an accuracy of 96.1% with the top 9 ranking genes. When using the lymphoma data set, the FGF performed best on the SVM classifier, achieving an accuracy of 100% with the top 12 ranking genes. The performance of the FGF is attributed to the fact that it is optimised to rank genes in such a way that results in maximum class separability, as well as its incorporation of multiple features of the data when ranking genes.

## 1. Introduction

Microarrays have revolutionised the way we analyse genomic composition and expression by allowing for high-throughput analysis of a tissue's genome and transcriptome [2]. Microarrays are used to quantify tissue mRNA content, allowing one to identify over or under expressed genes. A microarray consists of thousands of oligonucleotide probe-sets bound on a chip substrate and, for a given tissue sample, generates an expression value for each gene represented on the chip [2]. Microarray data can be used to identify malfunction in genetic expression and can also be used to develop diagnostic and prognostic classifiers [3-7].

One of the most important aspects of microarray data analysis is the identification of genes which characterise types and sub-types of cancers [8]. The expression values of these genes form the feature set used to implement cancer microarray classification systems. If good features are used for classification then even the simplest classifier can achieve high accuracies [9]. These features should demonstrate significant variation between samples of different types [8]. They are identified by first ranking the genes in order of differential expression, followed by a cross-validation scheme, which is used to identify the optimal number of top ranking genes required for classifier training and testing [10].

The Fuzzy Gene Filter (FGF), first described the author [1, 11], is a novel gene ranking algorithm, based on an adaptive Fuzzy Inference System. The FGF includes both parametric and non-parametric inputs and is optimised using a Genetic Algorithm (GA). Previously [1], the FGF was compared to three of the most common approaches to microarray feature ranking namely the t-test, Wilcoxon test and ROC curve analysis, based on the classification accuracy obtained using a K Nearest Neighbour classifier. In this paper a more thorough comparison is carried out, using four supervised classifiers to compare the feature ranking algorithms: Artificial Neural Network (ANN), Support Vector Machine (SVM), Naïve Baysian Classifier (NBC) and K Nearest Neighbour (KNN) classifier.

Two publicly available data-sets are used for training and testing. Results are also compared to those previously obtained using the same data-sets, as recorded in the literature.

This paper comprises five main sections. In the next section, the three most common approaches to gene ranking, as well as the FGF, are discussed. Section 3 describes the classifiers used to test the various ranking algorithms. The experimental design is described in section 4 and in section 5 test results are presented and discussed.

## 2. Gene Ranking

A typical expression array experiment produces data which consists of thousands of expression values per sample processed. Most microarray data analysis packages implement five distinct steps in microarray data analysis [4, 12, 13]:
1. Data pre-processing (intra-chip and inter-chip normalisation).
2. Gene Selection (identifying differentially expressed genes).
3. Clustering (identifying common expression patterns – co-expression analysis).
4. Functional Enrichment/Biological Pathway analysis (identifying the biological significance of the selected genes).
5. Classification (developing a classification system for unclassified samples).

Gene selection is the most central step in microarray data analysis [12, 14]. Genes are typically ranked in order of differential expression and the top ranking genes are selected for classifier training and functional enrichment analysis [10]. The three most common feature ranking approaches used are the Student t-test, Wilcoxon test and ROC curve analysis.

### 2.1 Student T-Test

The Student t-test was first proposed by William Sealy Gosset (who published under the pen name 'Student') in 1908 [15]. The two-sample t-test is a parametric hypothesis test which examines whether two data-sets were sampled from the same distribution (or have the same mean).

In the context of differential expression analysis, it is assumed that, for a particular gene, the expression values across two classes are of an unequal sample size and have an unequal variance [12, 16]. Hence an unpaired t-test is generally implemented on expression array data.

The t-statistic is directly proportional to the inter-class mean difference and inversely proportional to the intra-class standard deviations. Small intra-class standard deviations and a large inter-class mean difference is indicative of a good class differentiating gene (small p-value). A p-value is determined based on the overlap of the distributions. If the p-value is less than an arbitrary assigned p-value cut-off (defined by the required confidence interval) then the gene is classified as being differentially expressed.

### 2.2 Wilcoxon Test

Non-parametric techniques, such as the Wilcoxon test [17], have also been used on microarray data [16]. The Wilcoxon test, first advanced by Frank Wilcoxon in 1945, is a non-parametric hypothesis test which sums the ranks of samples of a particular class and based on the rank sum, determines a p-value.

In the context of microarray data, the expression values of a gene are ranked in ascending order [16]. The class distribution of the ranks is examined and the expression rankings of samples from the same class are then summed. If the sums of ranks for both classes are similar then the gene is not able to differentiate the classes, resulting in a large p-value. If the sum of ranks of the classes differ then the gene is differentially expressed and will have a small p-value.

### 2.3 ROC Curve Analysis

ROC analysis was originally used in signal detection theory to assess the accuracy of correctly classifying radar signals [18]. It has also been extensively applied to medical diagnostic performance analysis. ROC analysis has been recently applied to microarray gene ranking, where each gene is assigned a p-value based on its performance as a classifier [19]. In the context of Machine Learning, it has been used for model comparison by assessing the area under the ROC curve (AUC) and hence deriving the ROC AUC statistic [20]. This approach has been criticised since AUC is a noisy classification measure and has been proven to be problematic in model selection [20].

An improvement of the ROC AUC statistic involves evaluating the area between the classifier's ROC curve and the non-discriminatory line or random classifier slope [20]. Based on this area, a p-value is generated, evaluating the class distinctive performance of the classifier as compared to randomly guessing the class distribution. If the number of correctly guessed samples is the same the number of false alarms then the classifier is no better than randomly assigning class labels to each sample. This is represented by the random classifier slope. Hence, the area between the ROC curve and the random classifier slope evaluates the randomness associated with the classifier.

If the area is large, then the classifier demonstrates a high positive hit rate and a low false alarm rate, indicative of a good classifier, also demonstrating a low level of randomness. This will result in the classifier being assigned a small p-value. If, however, the area is small then the classifier is no better than randomly assigning class labels, resulting in a small p-value. In the context of expression data, each gene is treated as a potential classifier and the p-value calculated by evaluating the area between the classifier's ROC curve and the random classifier slope is used to rank the genes in order of class differentiability [19].

### 2.2 Fuzzy Gene Filter

The FGF is a rule based gene ranking tool utilising a Fuzzy Inference System [1, 11]. A Fuzzy Inference System [21-23] is a robust decisive tool which mimics the way human beings make decisions based on imprecise data.

At the core Fuzzy Inference is fuzzy set theory. Fuzzy set theory, as opposed to classic set theory, assigns each variable a degree of membership [21]: whereas Boolean logic only deals with binary membership, fuzzy logic can assign a single point to multiple groups with varying degrees of membership.

The motivation for using fuzzy logic for gene ranking lies in its ability to tolerate imprecise data [24]. Fuzzy logic is suitable for microarray data analysis due to its inherent imprecision - expression variation of biological replicates is inevitable. Also, due to the FGF's heuristic nature, diverse biological and statistical expert knowledge can be incorporated when ranking genes [24]. A schematic overview of the FGF is presented in Figure 1.

The FGF is based on a Mamdani fuzzy inference architecture (due to its intuitive implementation) and consists of four components: Input layer, input fuzzy membership functions, rule block and output fuzzy membership functions.

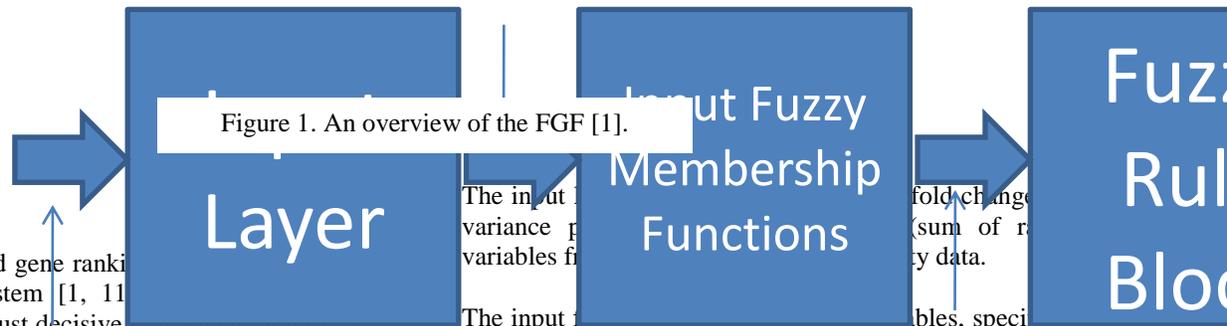

Figure 1. An overview of the FGF [1].

The input layer accepts the fold change, varianceptot and rank sum of rank variables from the microarray data.

The input fuzzy system has two variables, specifying the fuzzy regions, as depicted in Figure 2 ($\alpha$ and $\beta$ specify the width of the fuzzy region). The fuzzy region is optimised using a Genetic Algorithm (GA) by determining the optimal $\alpha$ and $\beta$ values, as depicted on Figure 3. A Separability Index (SI) is used to guide the GA toward the optimal fuzzy region.

Fuzzy rules are extracted from underlying statistics (both parametric and non-parametric). For example, if a gene has low intra-class variance, a high fold change and a high rank sum then the gene is deemed to display good class differentiability and is hence assigned to the very high output fuzzy membership function. For a full description of the FGF, the reader is referred to authors previous publications on the topic [1, 11].

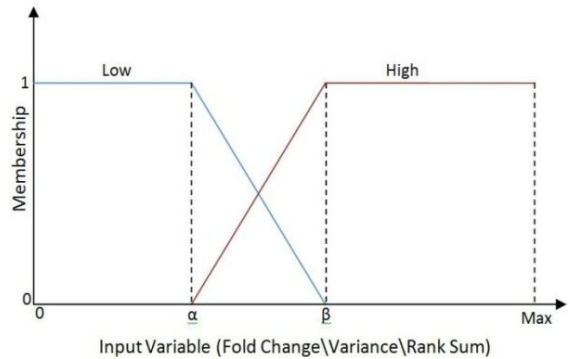

Figure 2. The FGF input membership functions [1].

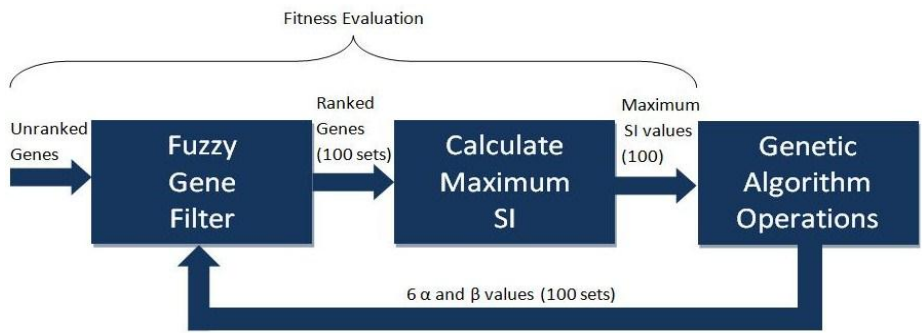

Figure 3. The FGF parameter optimisation scheme: A GA is used to optimise the fuzzy region of each variable [1].

# 4. Classifiers

The application of supervised classification systems for cancer diagnosis using microarray data has become prevalent and most microarray studies incorporate supervised classification as an indication of diagnostic feasibility [5, 10, 11, 25]. A number of studies have shown relatively high classification accuracies on types and subtypes of cancer samples ranging from lymphomas [26] to prostate cancer [27].

A wide range of supervised learning algorithms have been applied to microarray data for sample classification. Techniques ranging from Artificial Neural Networks (ANN) [28] to Support Vector Machines (SVM) [29], from K Nearest Neighbour (KNN) [30] to Naïve Bayesian Classifiers (NBC) [31] have been applied to microarray data classification and their adequacy assessed. One of the most thorough studies on the subject was carried out by Statnikov et. al. [10] who conclude that the best classifier architecture for cancer expression classification is the Support Vector Machine.

## 4.1 Artificial Neural Network (Multilayered Perceptron)

A neural network is a powerful modelling and prediction tool, used to model complex, multi-input, multi-output (MIMO) systems, as well as non-linear systems, by utilizing the input\output (I\O) data acquired from the system [32]. Neural networks stemmed from the desire to get machines to imitate the way the brain acquires knowledge, and hence gave rise to artificial intelligence. This is desired since, like the brain, neural network models have a very high degree of fault tolerance and parallel computing power. For example the brain is capable of dealing with multiple inputs and multiple outputs with ease as well as dealing with poor quality (noisy) signals. For example the brain can reconstruct and recognise a partial or poor quality visual.

There are various types of Artificial Neural Networks. The most common ANN is the Multilayered Perceptron [32](MLP) Neural Network (Figure 4).

A MLP [32] consists of a complex network of neurons. Neurons form connections between nodes. Each neuron stores knowledge in the form of a connection strength known as a weight. A weight describes the affect a particular node has on the node to which it is connected. A node stores knowledge in the form of a bias (a value added to the inputs at the node).

All the inputs to a node are summed and transformed to the output via an activation function (symbolised by the F blocks in Figure 4). The MLP is trained via back-propagation by presenting it a portion of the input data and comparing the outputs of the network to the target outputs, iteratively adjusting the weights and biases until the MLP's outputs approximate the targeted outputs [32]. A number of optimisation algorithms can be used to optimise the weights of the MLP, the most efficient being Scaled Conjugate Gradient (SCG) [32].

## 4.2 Support Vector Machine

Support Vector Machines (SVM), originally developed by Cortes and Vapnik et.al. [33] in the mid 1990's, are hard, non-parametric, robust classifiers, normally trained using supervised learning. The Support Vector Machine (SVM) is considered to be one of the most significant developments in Artificial Intelligent classification in recent years. The SVM's insensitivity to a high dimensional input space makes it an ideal candidate for classification of GEP.

SVMs operate in vector space. The classified input data is vectorised, as depicted in the simplified 2 dimensional vector space in Figure 5.

During training, a discriminant function, or decision boundary, is generated to separate between the two classes [33]. A margin between the discriminatory function and the nearest data points or vectors is then generated, as depicted in Figure 5. The vectors which result in the largest margin are referred to as support vectors.

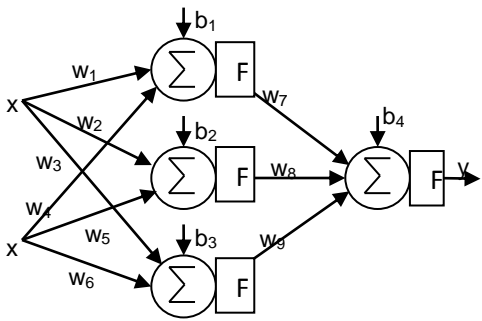

Figure 4. A Three layered Multilayer Perceptron.

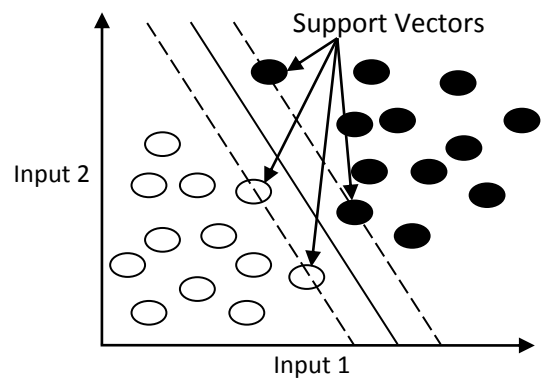

Figure 5. SVM Training.

Support vectors are identified through quadratic programming and Lagrange Multipliers. During cross-validation, the extent of influence of outliers is determined [33]. Outliers can result in a decision boundary with a smaller margin, resulting in suboptimal classification accuracy. Therefore, an upper-bound constant, normally symbolised by C, is defined in order to limit the influence of outliers.

A discriminant function is also known as a kernel. The type of kernel depicted in Figure 5 is a linear kernel. Other types if kernels include polynomial kernels and Gaussian kernels. Generally, for high dimensional vector spaces, linear kernels can achieve just as good accuracies as polynomial and Gaussian kernels.

If classes are not linearly separable, then the data is projected into a higher dimensional space where the classes can be separated using a linear hyperplane [33]. Furthermore, SVMs can be designed to be fairly robust towards outliers by setting the trade-off and penalty parameters.

### 4.3 K Nearest Neighbour

K-nearest neighbour (KNN) classification is a non-parametric classification technique first advanced by Cover et.al. [34] in the 1960's. KNN assigns an unknown sample to the class belonging to the majority of samples in it's neighbourhood.

The neighbourhood radius is specified by the number of nearest samples required to make a class assignment. The optimal neighbourhood radius is either pre-defined (for unsupervised learning) or learned during cross-validation. KNN is the simplest of the algorithms described in this chapter (and hence the least computationally expensive) yet has performed well on microarray data.

### 4.4 Naive Bayesian Classifier

The Naïve Bayesian Classifier (NBC) is a probabilistic classifier based on Bayes theorem and assumes that each feature is class-independent of one another. In the context of expression profiling, NBC assumes that each gene independently contributes to the probability that a sample belonging to a particular class [31].

A NBC, like any supervised classifier, undergoes training in order to establish the optimal parameters of the probability distribution [31]. Typically, a Gaussian distribution is assumed for each feature for each class and the optimal mean and standard deviation are identified during training. When classifying an unknown sample, The NBC calculates the posterior probability of the sample belonging to each class, by comparing the distributions of the samples features to those identified during training [31].

## 5. Experimental Design

The purpose of the experiment described in this paper is to examine how well the FGF performs in ranking features for various classification architectures (KNN, SVM, ANN, NBC), as compared to standard feature ranking approaches (t-test, Wilcoxon test, ROC curve analysis).

### 5.1 Data Sets

Two data-sets were used to facilitate the comparison, both made publically available by Statnikov [35]. The first consists of 50 healthy and 52 cancerous prostate samples. The second consists of 58 Diffuse Large B-Cell Lymphoma samples and 19 Follicular Lymphoma samples.

The prostate data-set was generated using the Affymetrix HG-U95 Gene Chip and consists of 10509 gene expression values per sample [27]. The lymphoma data-set was generated using the HU6800 oligonucleotide array and consists of 54070 gene expression values per sample [26]. Background correction was done using the Affymetrix MAS 5.0 algorithm. In addition, quantile normalisation with a median polish was also implemented.

### 5.2 Cross-validation

In order to assess the performance of various classifiers on features ranked by each of the ranking approaches, a cross-validation scheme is implemented in order to identify the optimal number of top ranking features to be used for classification. A classifier is iteratively re-trained and tested, incrementing the number of top ranking genes used until the gene-set which results in highest classification accuracy is identified. This gene-set is then selected as the classifier input space.

It is also necessary to identify the optimal classifier parameters, for each gene-set being tested. Hence, a nested stratified Leave-one-out Cross-validation (LOOCV) scheme is implemented [10]. The scheme consists of an inner loop and an outer loop. The inner loop identifies the optimal parameter values for the classifier (using a 10 fold cross-validation scheme). The outer loop calculates the LOOCV accuracy for the gene-set being tested.

LOOCV consists of training a classifier using all samples except for one. The classifier is then tested using the left-out sample. This process is repeated until each sample has been used to test the classifier. The LOOCV accuracy is then determined by calculating the percentage of correctly classified left-out samples.

LOOCV is commonly used for classification problems where there are a limited number of samples [10].

Table 1: Prostate data set classification accuracies and optimal number of top ranking genes.

|     | FGF | t-test | Wilcoxon test | ROC |
| --- | --- | --- | --- | --- |
| KNN | **96.1% (9)** | 93.1% (3) | 94.1% (15) | 93.1% (6) |
| SVM | **95.0% (3)** | 94.1% (14) | 94.1% (19) | 95.0% (8) |
| NBC | 94.1% (3) | 93.1% (22) | 93.1% (15) | **94.1% (3)** |
| ANN | **95.0% (5)** | 93.1% (7) | 94.1% (14) | 94.1% (6) |

Typically, one would allocate three sub datasets: A training set (used to train the classifier), a validation set (used to identify optimal classifier parameters and features) and a testing dataset (used to quantify the performance of the classifier on 'unseen' data). If there is a limited number of samples (relative to the number of features) then it is necessary to use the training dataset as the validation set as well and implement a cross-validation scheme, such as the one described here.

Due to the expense of generating microarray data, a typical microarray experiment consists of few samples, compared to the number of features generated per sample. Hence, LOOCV is common in microarray literature [10]. Furthermore, the purpose of this experiment is to compare feature ranking algorithms, as opposed to classifiers, hence the LOOCV accuracy is sufficient to compare feature sets.

This approach is also used since it is similar to the one used by the original authors of the test data-sets, where a KNN classifier was used to diagnose prostate cancer and differentiate between Diffuse Large B-cell Lymphoma and Follicular Lymphoma [26]. It is also similar to approach taken by Statnikov et.al. in a paper which compares various classifier architectures on microarray data [10].

In the inner loop of the LOOCV, the optimal parameters of the classifiers are identified. For each of the four classifiers tested, the following classifier parameters are optimised:
- For the SVM, the upper-bound constant C is optimised, while using a linear kernel as suggested by Statnikov et. al..
- For the KNN classifier the optimal neighbourhood radius k is identified.
- For the NBC the bandwidth of the initial Gaussian kernel is optimised.
- For the ANN (MLP), the optimal number of hidden nodes is identified (within the range of one hidden node to twice the number of input nodes) while using regularisation to prevent over-fitting. A logistic activation function is used since the MLP is being used as a classifier.

Once LOOCV has been implemented using each of the classification algorithms, on features ranked by each of the feature ranking techniques, classification accuracies are compared. An ANOVA is then implemented to examine the significance of the different accuracies across the various gene ranking algorithms.

### 5.3 Software

All techniques were implemented in MATLAB R2010a. The Bioinformatics toolbox was used to implement the t-test, the Wilcoxen test, the ROC test, SVM, KNN and NBC. The Fuzzy logic toolbox was used to implement the FGF. The NETLAB neural network toolbox was used to implement the MLP ANN. SI and cross-validation were coded in MATLAB.

## 6. Results and Discussion

Tables 1 and 2 depict the highest LOOCV accuracies attained by each classifier for each feature ranking algorithm, as well as the optimal number of top ranking genes used to attain the accuracy (the value in parenthesis). The same results were achieved every time the experiment was repeated.

The LOOCV accuracies attained for the various classifiers tested on the prostate data set are summarised on Table 1. Figure 6 depicts the butterfly diagrams of the various classifiers, depicting the accuracy median and $25^{th}$\$75th$ percentiles of the four classifiers.

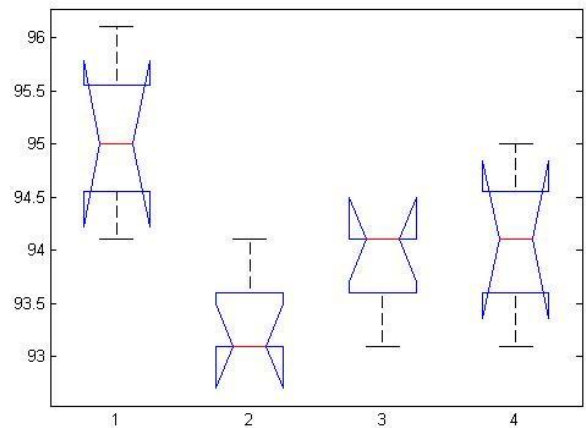

Figure 6. Butterfly diagram for the prostate data set, displaying the distribution of classification accuracies for each feature ranking technique (sample 1 is the FGF, 2 the t-test, 3 the Wilcoxon test and 4 ROC curve analysis).

Table 1: Prostate data set classification accuracies and optimal number of top ranking genes.

|  | FGF | t-test | Wilcoxon test | ROC |
|---|---|---|---|---|
| KNN | **100% (13)** | 97.4% (6) | 94.8% (4) | 98.7% (2) |
| SVM | **100% (12)** | 98.7% (5) | 98.7% (39) | 98.7% (28) |
| NBC | **97.4% (5)** | **97.4% (3)** | **97.4% (5)** | **97.4% (3)** |
| ANN | **98.7% (14)** | 94.8% (8) | 97.4% (6) | 97.4% (4) |

Classifiers trained with features ranked by the FGF resulted in the highest accuracy, for each of the classifiers tested, compared to the other gene ranking techniques (p < 0.0231). The classifier with the highest accuracy is the KNN classifier, attaining an accuracy of 96.1%, when trained using the top 9 ranking genes, as ranked by the FGF. The classifier with the highest accuracy is the KNN classifier, attaining an accuracy of 96.1%, when trained using the top 9 ranking genes, as ranked by the FGF.

The prostate data-set was originally used by Singh et. al. [27] to develop a classifier for prostate cancer diagnosis. The maximum cross-validation accuracy reported in the original paper was 86% using a 16 gene model (genes were ranked using a signal to noise ranking scheme). Statnikov et. al. [10] reported an accuracy of 92% on the same data-set. All the gene ranking techniques presented here outperformed both studies with the FGF attaining an accuracy of 96.1% using a KNN classifier.

The LOOCV accuracies attained for the various classifiers tested on the lymphoma data set are summarised on Table 2. Figure 7 depicts the butterfly diagrams of the various classifiers, depicting the accuracy median and $25^{th}$\\$75^{th}$ percentiles of the four classifiers.

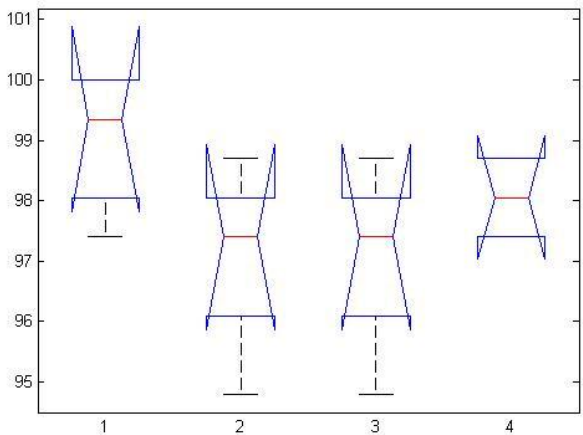

Figure 6. Butterfly diagram for the prostate data set, displaying the distribution of classification accuracies for each feature ranking technique (sample 1 is the FGF, 2 the t-test, 3 the Wilcoxon test and 4 ROC curve analysis).

Classifiers trained with features ranked by the FGF resulted in the highest accuracy, for each of the classifiers tested (p < 0.1888), albeit with less confidence that than with the prostate data set. Both the KNN and SVM classifiers attained the highest accuracy (100%). Nevertheless, the SVM is deemed the better classifier since it was able achieve the maximum accuracy with fewer features (the top 12 ranking genes as opposed to the top 13 with the KNN classifier). Similarly, even though the features ranked by ROC curve analysis also resulted in 100% accuracy on the SVM, it did so with the top 28 features. The features ranked with the FGF achieved the same accuracy with only 12 top ranking features.

The lymphoma data-set was originally used by Shipp et. al. [26]. The accuracy reported in the original paper was 77% using weighted voting classification technique. Statnikov et. al. [10] reported an accuracy of 97.5% on the same data-set. The FGF outperformed both studies, attaining an accuracy of 100% using the SVM classifier.

The performance of the FGF is attributed to the fact that it is optimised to rank genes in such a way that results in maximum class separability, as well as its incorporation of multiple features of the data when ranking genes.

Furthermore, the FGF parameters are optimised to the specific data-set being analysed: the optimised fuzzy parameters for the prostate data-set are different to those of the lymphoma data-set, as discussed by Perez et. al: the FGF α and β values for the fold change membership functions, optimised for the prostate data-set, are 0.0862 and 0.7787. In contrast, the FGF α and β values, optimised for the lymphoma data-set, are 0.1098 and 0.5378. This is indicative of the difference in the diseases being tested, as described by Perez et. al. [1].

## 7. Conclusion

After doing a thorough comparison of the FGF with standard gene ranking algorithms (the t-test, Wilcoxon test and ROC curve analysis), on various classifier architectures (KNN, SVM, NBC and ANN), the FGF was still able to attain the highest LOOCV accuracy on both data-sets (p < 0.0231 for the prostate data set and p < 0.1888 for the lymphoma data set).

For the prostate data set, a LOOCV accuracy of 96.1%, using the top 9 ranking genes, was attained the KNN classifier. For the lymphoma data set, a LOOCV accuracy of 100%, using the top 12 ranking genes, was attained on the SVM classifier. It is thus evident that (at least for the data sets tested) ranking genes using the FGF results in the section of a better feature set than when ranked with standard approaches, no matter which classifier is used for classification. This is due to the fact that the FGF is optimised for the specific data set being analysed and due to its incorporation of both parametric and non-parametric features for ranking genes.